# Transformer-F: A Transformer network with effective methods for learning universal sentence representation

**Yu Shi**[1,*]

[1]School of Software Engineering, Beijing University of Posts and Telecommunications, Beijing, China
[1]Key Laboratory of Trustworthy Distributed Computing and Service, Beijing University of Posts and Telecommunications, Beijing, China
{yus@bupt.edu.cn}

## Abstract

The Transformer model is widely used in natural language processing for sentence representation. However, the previous Transformer-based models focus on function words that have limited meaning in most cases and could merely extract high-level semantic abstraction features. In this paper, two approaches are introduced to improve the performance of Transformers. We calculated the attention score by multiplying the part-of-speech weight vector with the correlation coefficient, which helps extract the words with more practical meaning. The weight vector is obtained by the input text sequence based on the importance of the part-of-speech. Furthermore, we fuse the features of each layer to make the sentence representation results more comprehensive and accurate. In experiments, we demonstrate the effectiveness of our model Transformer-F on three standard text classification datasets. Experimental results show that our proposed model significantly boosts the performance of text classification as compared to the baseline model. Specifically, we obtain a 5.28% relative improvement over the vanilla Transformer on the simple tasks.

## 1 Introduction

Learning sentence representation is the basis for many natural language processing tasks, which aims to learn a fixed-length feature vector that contains both the semantic and syntactic information of the sentences. Much progress has been made in learning semantically meaningful distributed representations of sentences. The current mainstream methods for learning sentence representations are based on deep learning and neural networks. The popularly used sentence representation models are the following three types: CNN-based models, RNN-based models, and Transformer-based models.

Figure 1: Examples of BERT as the pre-processor for classifying. Blue indicates feature importance represents the extent that the word was essential towards the classifier.

The Transformer-based models have more parallelization capability than RNN and solve the problem that CNN cannot capture long-distance dependent. In contrast, it allocates the same computational resources for different words, prone to waste of resource wastage. And it is limited by a fixed-length context in the setting of language modeling, especially on long sequences where the cost of training these models can be prohibitive. To improve the Transformer and circumvent the shortcomings of the vanilla model, some researchers proposed different variants of the Transformer, such as Universal Transformer (Dehghani et al., 2019), Transformer-XL (Dai et al., 2019), Reformer (Kitaev et al., 2020), and Linformer (Wang et al., 2020). Additionally, Wang et al. (2020) provide an in-depth consideration of the Transformer architecture. They have evaluated the importance of each module in the model and then pruning or parameter rewinding the unimportant modules so that the model can be optimized.

---
[*] Corresponding author.

Nevertheless, Transformer can only extract high-level semantic abstraction features, lack the use of shallow feature information, which affects the performance of simple tasks. In addition, the previous Transformer-based models focus on function words that have little meaning in many cases. Figure 1 displays examples of BERT as the generator for classifying. It is evident that self-attention is concerned about some of the function words, such as "的", "了" and "么". To address these issues, we introduce a sentence representation model Transformer-F, which can aggregate notional words for sentence representation and fuses characteristics from each layer aiming to improve model performance. Explicitly, we have the following contributions:

- The correlation coefficient is used instead of dot-product for extraction of specific components of the sentence.

- We multiply the part-of-speech weight vectors with the correlation coefficients matrix to calculate the attention weights, which help select words with more practical meanings. Concretely, we convert the input text sequence into the corresponding part-of-speech weight sequence depending on the importance of part-of-speech. To accelerate the extraction of part-of-speech sequence, we compile a comparison table of commonly Chinese lexical parts of speech using existing Xinhua dictionary data combined with the common standardized Chinese character list.

- Fusing the features of each layer makes the model's sentence representation results more comprehensive and accurate, thus improving the performance of the model.

We trained five different models on three standard text classification datasets to evaluate Transformer-F's performance. And then, we evaluated the feasibility of our strategies on CED datasets from the dimensions of Accuracy and F1. It shows that the Transformer-F we trained has superior sentence representation and outperforms the standard Transformer on text classification. Especially on small datasets, our model has significant improvement.

## 2 Related Work

In this section, we introduce the relevant background on the generation of universal sentence representation and Transformer-based models.

### 2.1 Sentence Representation

Text is a prosperous source of information. However, its unstructured property leads to the sentence representation of text become a challenging task. Deep learning and neural networks are currently dominant sentence representation methods for text classification tasks and have been explored to address the limitations of traditional representation. In this section, we focus on the research progress of deep learning in sentence representation.

The CNN-based models retain natural language relative location information and have a high degree of parallel freedom, but they cannot capture long-range features. In order to obtain semantic information at different levels of abstraction, Kim (2014) applied CNN for text classification to achieve the combination and filtering of N-gram features. Zhang et al. (2016) treat character-level text as the original signal and uses a one-dimensional convolutional neural network for feature extraction, which can effectively alleviate the Out of Vocabulary (OOV) problem effectively learning the feature representation. However, the CNN-based feature representation model suffers from the inability to capture long-distance features and does not adapt to variable-length text sequences, which can impair the performance of text classification tasks to some extent.

The RNN-based models treat text as sequences of words and aim to capture the correlation between words and information about text structure, which solves the long-distance dependency problem to a certain extent. Moreover, it can handle variable length sequences. But the sequence-dependent structure of RNNs makes it challenging to have efficient parallel computing capabilities. Liu et al. (2016) applied RNN to text classification and proposed a multi-task structure based on RNN to improve the recognition effect of the model. Lai et al. (2015) combined RNN and CNN to construct a new model that exploits both models' advantages and is less noisy than the traditional window-based neural network. This model maximizes the extraction of contextual information and automatically decides which features play a more critical role. However, RNN and its variants are hardly capable of efficient parallel computation, which is fatal for big data processing.

What is more, some people propose to use attention mechanism on top of the CNN or RNN model to introduce information to guide the extraction of sentence representation. Yang et al. (2016) proposed a hierarchical attention network HAN for text classification, which has a significant advantage over previous methods on six text classification tasks. This model has two distinctive characteristics: (i) the sentence-level attention model learns which sentences in the document are more important for determining overall sentiment, and (ii) the word-level attention model learns which words in each sentence are decisive. ATAE-LSTM (Zhou et al., 2016) extends the hierarchical attention model to cross-language sentiment classification. In each language, the LSTM network is applied for model documents, and the hierarchical attention mechanism is used for classification. In each language, the documents are modeled with LSTM networks. And then, classification is achieved by using a hierarchical attention mechanism. Those models created new state-of-the-art on text classification tasks when they were published.

### 2.2 Transformer

Attention mechanisms enable a neural network to focus more on relevant elements of the input than on irrelevant parts. In a milestone paper (Vaswani et al., 2017), the Transformer attention module is presented, superseding past works and substantially surpassing their performance. Since its introduction, Transformer as a particular attention-based model that has created a new baseline for deep learning because of its excellent sentence representation capabilities. It enables parallelized computation and captures long-distance dependent feature problems, and it is the most popular model for textual sentence representation.

To circumvent the shortcomings of the standard Transformer, many researchers proposed different variants. Universal Transformer addresses the shortcomings of vanilla Transformer's Non-Turing completeness and the problem of averaging the computational resources devoted, thus improves Transformer's performance on simple tasks. Transformer-XL combines the advantages of RNN sequence modeling and self-attention mechanism. It uses a multi-head attention module on each segment of the input sequence and a loop mechanism to learn the dependencies between consecutive segments, which solving the long-distance dependency problem without a significant increase in arithmetic power consumption. Reformer proposes a locally sensitive hash to replace the original dot product approach of attention to reduce the model complexity. At the same time, the inverse residual layer is also used to replace the standard residuals to reduce the model memory footprint. It improves the memory efficiency and computing speed of the model while maintaining a comparable performance with Transformer. Linformer reduces the overall self-attention complexity from $O(n2)$ to $O(n)$ in both time and space and performs on par with standard Transformer models, which demonstrate that a low-rank matrix can approximate the self-attention mechanism.

Whereas these methods have made some new findings, Transformer remains to focus on function words with little meaning in many cases. How to fuse each layer of neural network information to improve model feature representation is also a pressing issue. For these problems, we propose Transformer-F, which can capture the semantic and syntactic information of sentences and obtain an accurate and comprehensive representation of sentences. Our solid experimental results show that Transformer-F achieves desirable results on several publicly available datasets.

### 3 Methodology

In this section, we present our proposed model in detail. First of all, we use correlation coefficient instead of dot-product to represent the relevance between words properly. Next, we convert the input text sequence into the corresponding part-of-speech weight vector in order for the model to extract the notional word, and the final attention is calculated by multiplying the part-of-speech weight sequence with the correlation coefficient matrix. And then, we fuse the features extracted from each layer to enable the model to extract more comprehensive and accurate features to improve the model performance. The model overview is shown in Figure 2.

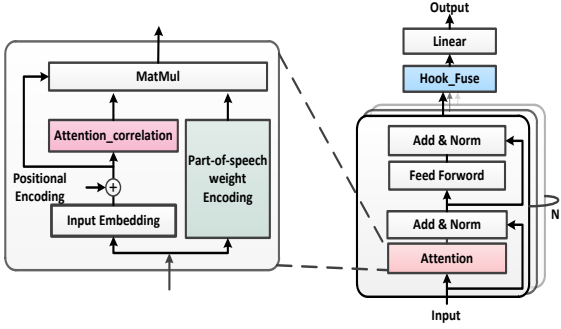

Figure 2: Overview of the Transformer-F.

### 3.1 Correlation Attention

We first rewrite the normal attention to formalize the attention equation we use in this paper:

$$ATT\_COR(X) = SOFTMAX(RX) \quad (1)$$

To find the relationship between the sequences internally, we try to introduce the correlation coefficient matrix instead of the dot-product. Since in statistics, correlation coefficients are often used to reflect the degree of correlation between variables. This approach makes better use of the features extracted by the model and reflects the correlation between words more precisely. Usually, the correlation coefficient matrix $R$ is expressed as:

$$R = \begin{pmatrix} \rho_{11} & \rho_{12} & \cdots & \rho_{1n} \\ \rho_{21} & \rho_{22} & & \rho_{2n} \\ \vdots & & \ddots & \vdots \\ \rho_{n1} & \rho_{n2} & \cdots & \rho_{nn} \end{pmatrix} \quad (2)$$

The correlation coefficient between the columns of the matrix $\rho_{ij}$ is:

$$\rho_{ij} = cov(X_i, X_j) / (\sqrt{DX_i}\sqrt{DX_j}) \quad (3)$$

$X_i, X_j$ is the vector standing for a $d$ dimentional word embedding for the $i$-th word in the sentence. $DX_i$ and $DX_j$ are the variances of $X_i$ and $X_j$ respectively. Using Expectation $E$ to calculate covariance $cov(X_i, X_j)$, which is shown as:

$$cov(X_i, X_j) = E((X_i - E(X_i)) \cdot (X_j - E(X_j))) \quad (4)$$

In addition, we solve the problem of an extensive range of data attribute deviations of the correlation coefficient matrix by regularizing the correlation coefficient matrix. This method is the process of scaling individual samples to have a unit norm, and it can be helpful to quantify the similarity of any pair of samples. However, the final matrix of correlation coefficients obtained is quite dense, which is a considerable burden. Also, even if the attention scores are high, it does not mean that this is critical for the words to simulate the decisions.

### 3.2 POS Feature extraction

The final decision of the model does not rely solely on the word-to-word interaction information learned by the model. Through experimental analysis, we noticed that in some cases: many of the word tags with large attentional weights were adjectives or adverbs that conveyed explicit signals on the underlying class tags. We argue that the model should be given word features that convey clear signals, making them more likely to receive greater attentional weight in a given sentence. Therefore, we propose to introduce part-of-speech annotation information during feature extraction.

We have compiled a comparison table of commonly used Chinese lexical parts of speech using existing Xinhua dictionary data combined with the typical standardized Chinese character list. The table contains 8105 words: the first level word list is a set of commonly used words, 3500 words, mainly to meet the basic needs of primary education and cultural popularization with words; The second-level character list contains 3,000 characters and is second only to the first-level characters in terms of usage; The word list of the third level includes 1605 words, which are the more common words in the family names, geographical names, scientific and technical terms and the words used in the language textbooks of primary and secondary schools that are not included in the first and second level word lists, mainly to meet the needs of words used in specialized fields closely related to public life in the information age.

$$ATTENTION(l) = ATT\_COR(X) \cdot W\_POS \quad (5)$$

We combine the obtained correlation coefficient $ATT\_COR$ with the input part-of-speech tagging weight vector $W\_POS$ as the attention score $ATTENTION$ for the final feature extraction shown as Equation 5, which helps extract the words with more practical meaning.

| Data set | language | classes | source | Number of data | Average #w | vocabulary |
|---|---|---|---|---|---|---|
| THUCNews | Chinese | 10 | RSS of Sina News | 200,000 | 22 | 4803 |
| CED | Chinese | 2 | Rumor data of Sina | 3,387 | 113 | 4411 |
| MR | English | 2 | Movie Reviews | 10,662 | 116 | 18779 |

Table 1: Data statistic: #w denotes the number of words per sentence.

However, we find that the intention of Chinese character symbols shifts over time, and knowledge can be transferred from the dataset to the target task in many different ways, so we optimize Transformer-F using a part-of-speech tagging interface for periodically expanding and updating the comparison table.

### 3.3 Hook Fusion

Transformer usually takes the output of the last encoder of the neural network as the feature representation of the text. Nevertheless, this representation may be too coarse to describe the features on the local space accurately, and the features extracted by the first encoder block of Transformer are too shallow to represent high abstract semantic information. Inspired by Wang and Tu. (2020), we choose to fuse the output features of each layer. The fused features have both higher semantics and resolution, which facilitates simple tasks and improves accuracy.

In this paper, we use the hook function as an operator to fuse the features extracted from each layer for the purpose of obtaining infinitely close to the optimal extracted features as the final result of the text feature representation. The hook function can be written as:

$$Hook(l) = (al + b/l)/(N/2) \ (ab > 0) \quad (6)$$

N is the number of encoder blocks, $l$ indicates how many Encoder blocks it is, a and b are hyperparameters of hook function. We combine the features representation from each encoder block using the hook function to obtain the final output as:

$$output = Linear(\sum_{l=0}^{N} Hook(l)(x + Sublayer(x))) \quad (7)$$

We have an input sentence $S$ in Figure 1, which is mapped into a fixed-length vector with positional encoding as the final input, and then fed it into Transformer-F. We calculate the correlation coefficients between the words to get the correlation coefficient matrix $ATT\_COR(X)$. Furthermore, part-of-speech weight $W\_POS$ is obtained by using the comparison table of commonly used Chinese part-of-speech. We combine $ATT\_COR$ with $W\_POS$ as the final attention score $ATTENTION$ for extracting the words with more practical meaning. In the end, the output of each encoder block is fused using the hook function to obtain the final feature representation results.

We evaluated Transformer-F on three standard text classification datasets. It shows that Transformer-F we trained is better than the vanilla Transformer on text classification, and all the metrics are significantly improved. We will show the details of the experiments in the Experiments and analysis.

## 4 Experiments and analysis

In this section, we evaluate the effectiveness of our Transformer-F model on text classification tasks, mainly focusing on evaluating the quality of learned sentence representations.

### 4.1 Experimental Setting

Text classification is the most common and important task in NLP domain, so we choose to evaluate Transformer-F on the text classification task. The three standard datasets shown in Table 1 are used.

The selection of text features is the foundation and essential element of text mining and information retrieval. Deep learning-based feature extraction can automatically and quickly acquire new valid feature representations from the training data. We compare Transformer-F with several baseline methods, including DPCNN (Johnson and Zhang, 2017), Att-BLSTM (Zhou et al., 2016), Transformer, and Universal Transformer:

- DPCNN proposes a low-complexity word-level deep convolutional neural network architecture for text categorization that can efficiently represent long-range associations in text.

- Att-BLSTM means Attention-Based Bidirectional Long Short-Term Memory Networks to capture the most important semantic information in a sentence.

- Transformer has more parallelization capability than RNN and solves the problem that CNN cannot capture long-distance dependent.

- Universal Transformer (UT) is a parallel-in-time self-attentive recurrent sequence model that can be cast as a generalization of the Transformer model and improves accuracy on several tasks.

In this paper, we choose accuracy and F1 score as the evaluation metrics. Then, we processed the data with word-level and used the Chinese pre-trained word vector trained by Shen et al. (2018) on Chinese datasets. We trained 5 models in total, where the hyperparameters of the Transformer-F are as follows: the number of encoder blocks $N = 6$, the number of attention heads per layer $h = 1$, the size of hidden embeddings $d_{model} = 512$, the parameters of the hook function $a = 0.4, b = 2.9$ and the Adam algorithm with learning rate of 5e-4 was used for optimization. For the words that convey the practical meaning, we set $A\_weight = 1$ and for the rest of the words, we set $R\_weight = 0.5$.

### 4.2 Experimental Result

We trained five models on three text classification public datasets and compared Transformer-F with DPCNN, Att-BLSTM, Transformer and UT, and the accuracy results obtained from the experiments are shown in Table 2. It can be seen that the Transformer-F model proposed in this paper shows promising results in all three datasets, which are better than the vanilla Transformer.

These results suggest that Transformer-F improves 0.14% over vanilla Transformer on the THUCNews dataset for multi-classification tasks with the same parameters. It can be seen from the results that the effect of Transformer-F is close to that of DPCNN and ATT-LSTM. To make the experiments more explorative, we choose the MR dataset is small, which can easily lead to overfitting. To demonstrate the effect of Transformer-F directly and obviously, we chose to process the word-level data without pre-processing. The experimental results show that although Transformer-F improves the accuracy by 1.76% over Transformer, neither of them achieves the optimal results on the MR dataset. The average sentence length of MR is 116, which implies that Transformer-F captures poorer long-range dependencies than the RNN structure and Transformer-F can be improved further. In addition, Transformer-F achieves tremendous improvement on the CED_dataset, better than the other baseline models, and achieves optimal results with an accuracy of 82.11%, which is 5.28% better than the vanilla Transformer. These experimental results show that Transformer-F is more suitable for small sample binary classification tasks than the vanilla Transformer and can extract more accurate and comprehensive features for significantly improving task performance.

| Method | THUCNews | CED | MR |
|---|---|---|---|
| DPCNN | **90.98** | 81.82 | 68.54 |
| Att-BLSTM | 90.89 | 76.83 | **72.57** |
| Transformer | 89.56 | 76.83 | 61.33 |
| UT | 89.67 | 79.18 | 61.52 |
| Transformer-F | **89.70** | **82.11** | 63.09 |

Table 2: Accuracy of five models on three text classification datasets.

### 4.3 Ablation Experimental

Using accuracy alone is not a good measure of how well a sentence representation does when the data is unbalanced. Therefore, we chose to analyze the performance of the Transformer-F on the CED dataset from the dimensions of Accuracy and F1, which are commonly used evaluation metrics for text classification. We analyze the strategies mentioned on the CED dataset for better evaluation of the effectiveness of the strategies proposed in this paper: (i) Introducing only part-of-speech weight sequence in order for the model to extract notional word; (ii) Using only correlation coefficient instead of dot-product to represent the relevance between words properly; (iii) Using only the hook function as an operator to fuse the features

| Method | Acc | F1 |
|---|---|---|
| Transformer | 76.83 | 76.83 |
| +W_POS | 78.30 | 78.30 |
| +ATT_COR | 80.94 | 80.83 |
| +Hook_multi | 77.13 | 76.68 |
| Transformer-F | **82.11** | **81.99** |

Table 3: Accuracy and F1 score of each strategy on CED_dataset.

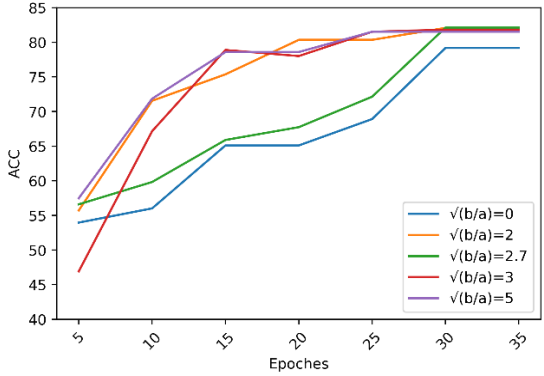

Figure 3: Accuracy curve for how the hook function parameters influence the performance of the CED. The horizontal axes indicate training epochs and the numbers in the legends stand for the corresponding values of $\sqrt{b/a}$.

extracted from each layer; (iv) The hybrid model Transformer-F.

The results show that our methods improve the results of the text classification task, as shown in Table 3. The attention score based on the correlation coefficient matrix has a noticeable performance improvement of 4.4% for the feature extractor. These results demonstrate that the correlation coefficient matrix is more effective than dot-product, which enhances model performance by preferably representing the word-to-word correlation. In addition, converting the input text sequence into the corresponding part-of-speech weight sequence in order for the model to extract notional words can also improve the model performance. The accuracy on CED_Dataset is about 1.76% better than that of the vanilla Transformer. Then, our proposed method of fusing the feature representations of the output of each layer improves the accuracy of the task by 0.59%. We combine these methods as our final model Transformer-F, which improves accuracy by 5.28% over the vanilla Transformer.

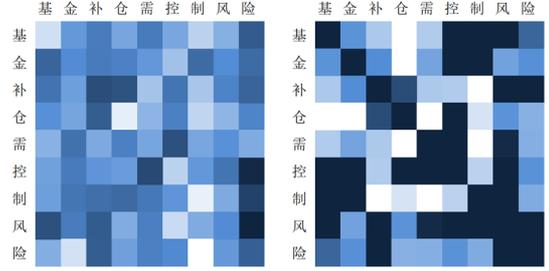

Figure 4: Self-attention weight VS ATT_COR weight. The color is darker for larger values. Transformer-F tends to identify more sparse word interactions that contribute to the final model decision.

To further analyze the effectiveness of our method, we conduct comparative experiments on the hyperparameters of the fuse method. Figure 3 describes the evaluation results of hook function parameters. We observe that our method leads the sentence representation more comprehensively, which significantly improves the accuracy of text classification. Especially, the result is optimal when the parameters of the hook function $\sqrt{b/a} = 2.7$, which also proves the lower encoder and upper encoder-attention are more critical.

In addition, we plot the heatmap to compare the self-attention weight with the ATT_COR weight proposed in this paper to analyze the difference between our self-attentive mechanism and the vanilla Transformer model, as shown in Figure 4 Transformer-F enhances the association between "基金" and "控制风险", which enables the model to extract the semantic representation of sentences better. It is demonstrated that the attention scores obtained using Transformer-F are more representative of the word interaction information between words with actual meaning, which contributes to the final decision of the model.

The above experimental results show that our proposed model can extract notional words with more practical meaning and makes full use of highly semantic abstracted features and shallow features, thus improving the performance of the sentence representation.

## 5 Conclusion

In this paper, we proposed Transformer-F for sentence representation to address the problem that vanilla Transformer focuses on function words that have little meaning and cannot utilize shallow features. We have improved the Transformer in three ways: (i) Using correlation coefficient instead of dot-product to represent the relevance between words properly; (ii) By multiplying the part-of-speech weight vectors with the correlation coefficients matrix we calculate the attention weights, which helps to extract words with notional words; (iii) Fusing the features of each layer to make the sentence representation results more comprehensive and accurate. We evaluated our model Transformer-F on three standard text classification datasets and verified the feasibility of the above strategies in turn. The experimental results show that sentence representation based on the Transformer-F model can significantly improve text classification, especially the 5.28% accuracy boost over the Transformer on the CED_dataset, which proves the availability of our innovation point. Although the Transformer-F model achieves good results, there is still much room for improvement. In future work, we will try to introduce the model to the long text feature representation task. In addition, this paper has only done experiments on text classification tasks, and we will try to test the performance of the Transformer-F model in machine translation, named entity recognition, and other tasks.

## Acknowledgments

We would like to thank the anonymous reviewers for their thoughtful and constructive comments. Any opinions, findings, and conclusions, or recommendations expressed in this material are those of the authors. The authors declare that there is no conflict of interest regarding the publication of this article.